\documentclass[letterpaper, 10 pt, conference]{ieeeconf}
\IEEEoverridecommandlockouts           

\usepackage{cite}
\usepackage{amsmath,amssymb,amsfonts}
\usepackage{graphicx}
\usepackage{textcomp}
\usepackage[dvipsnames]{xcolor}
\def\BibTeX{{\rm B\kern-.05em{\sc i\kern-.025em b}\kern-.08em
    T\kern-.1667em\lower.7ex\hbox{E}\kern-.125emX}}
    \usepackage{stackengine}
\usepackage[utf8]{inputenc} 
\usepackage[T1]{fontenc} 
\usepackage{graphicx}
\usepackage{float} 
\usepackage{caption}
\usepackage{url}
\usepackage{mathtools} 
\usepackage{mathrsfs}
\usepackage{colortbl} 
\usepackage{siunitx} 
\usepackage{fancyhdr}
\usepackage[makeroom]{cancel}

\let\labelindent\relax
\usepackage{amsthm}
\usepackage{subcaption}
\usepackage{tikz}
\usetikzlibrary{patterns,angles,quotes,arrows}
\usepackage{cite, enumitem}
\usepackage[normalem]{ulem}
\usepackage{booktabs}
\usepackage[belowskip=-15pt,aboveskip=0pt]{caption}
\usepackage{multirow}

\usepackage{listings}
\usepackage{hyperref}
\usepackage{bigints}
\usepackage{algorithm2e}
\usepackage{algpseudocode}

\usepackage{url}
\usepackage{hyperref}

\RestyleAlgo{ruled}
\SetKwComment{Comment}{/* }{ */}

\DeclareMathOperator*{\argmin}{arg\min}

\newtheorem{Problem}{Problem}

\newtheorem{Definition}{Definition}

\newcommand{\pg}{p^\mathrm{goal}}
\newcommand{\pgtemp}{\bar{p}^\mathrm{goal}}
\newcommand{\nrays}{n_\mathrm{rays}}

\newcommand{\norm}[1]{\left\lVert#1\right\rVert}

\begin{document}

\title{Foundation Models to the Rescue: Deadlock Resolution in Connected Multi-Robot Systems
\thanks{The authors are with the Department of Aeronautics and Astronautics at MIT, \texttt{\{kgarg, jarkin, szhang21, chuchu\}@mit.edu}. Project website: \href{https://mit-realm.github.io/LLM-gcbfplus-website/}{https://mit-realm.github.io/LLM-gcbfplus-website/}}
}
\author{Kunal Garg \and Songyuan Zhang \and Jacob Arkin \and Chuchu Fan}

\maketitle


\begin{abstract}
Connected multi-agent robotic systems (MRS) are prone to deadlocks in an obstacle environment where the robots can get stuck away from their desired locations under a smooth low-level control policy. Without an external intervention, often in terms of a high-level command, a low-level control policy cannot resolve such deadlocks. Utilizing the generalizability and low data requirements of foundation models, this paper explores the possibility of using text-based models, i.e., large language models (LLMs), and text-and-image-based models, i.e., vision-language models (VLMs), as high-level planners for deadlock resolution. We propose a hierarchical control framework where a foundation model-based high-level planner helps to resolve deadlocks by assigning a leader to the MRS along with a set of waypoints for the MRS leader. Then, a low-level distributed control policy based on graph neural networks is executed to safely follow these waypoints, thereby evading the deadlock. We conduct extensive experiments on various MRS environments using the best available pre-trained LLMs and VLMs. We compare their performance with a graph-based planner in terms of effectiveness in helping the MRS reach their target locations and computational time. Our results illustrate that, compared to grid-based planners, the foundation models perform better in terms of the goal-reaching rate and computational time for complex environments, which helps us conclude that foundation models can assist MRS operating in complex obstacle-cluttered environments to resolve deadlocks efficiently.
\end{abstract}

\section{Introduction}
Multi-agent robotic systems (MRS) are widely used in various applications today, such as warehouse operations \cite{baiyu2023DD,wurman2008coordinating}, self-driving cars \cite{dinneweth2022multi}, and coordinated drone navigation in a dense forest for search-and-rescue missions \cite{tian2020search}, among others. In various MRS applications for navigating in unknown environments, such as coverage \cite{cortes2004coverage} and formation control \cite{mehdifar2020prescribed}, {robot agents must remain within the communication region of each other, or, in other words, remain connected, so that they can actively communicate with each other to share information and build the unknown or partially known environment collectively}. Additionally, ensuring safety in terms of collision avoidance and scalability to large-scale multi-agent problems are also crucial requirements of the control design of MRS. When the requirements of connectivity and safety come together, existing methods for multi-agent coordination and motion planning often result in deadlocks, {where agents get stuck away from their desired goal locations}. Particularly, with the additional requirement of connectivity, even one robot getting stuck in a deadlock results in all the agents getting stuck, making it crucial to resolve the deadlocks for connected MRS. 

\begin{figure*}
    \centering
    \includegraphics[width=1\linewidth]{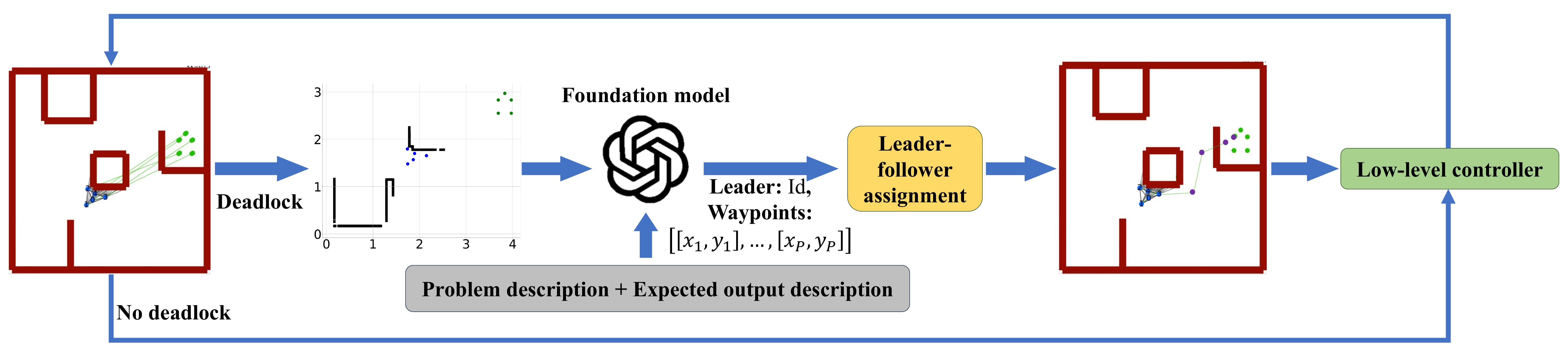}
    \caption{Overview of the hierarchical control framework using a foundation model as a high-level planner. In the image on the left, the robots are shown in blue, their goals in green, and obstacles in red. The planner assigns a leader and waypoints (shown in purple) for the leader of the MRS, resulting in a leader-follower formation. A GNN-based low-level controller provides a distributed control policy for safety and connectivity while the leader helps evade the deadlock.}
    \label{fig:algo overview}
\end{figure*}

In recent years, learning-based methods have shown promising results in computing a low-level control policy for complex robotic systems \cite{dawson2023safe,zhang2023distributed,garg2023learning}. The recent work \cite{zhang2024gcbf+} proposed a new notion termed graph control barrier function (GCBF) for encoding safety in arbitrarily large MRS, and a framework, GCBF+, for learning the GCBF and a safe distributed control policy. However, that work focuses on safety, i.e., collision avoidance, and does not incorporate connectivity maintenance or goal-reaching requirements. 
In this work, we modify the GCBF-based distributed low-level control policy so that the new policy can maintain both safety and connectivity for arbitrarily large MRS. 
However, the resulting low-level control policy is still prone to failure modes such as deadlocks in obstacle environments. Many works have been proposed to solve the problem of detecting and moving out of deadlocks under safety constraints \cite{zhou2020distributed,grover2021deadlock,grover2023before,chen2024deadlock}. However, these works do not consider connectivity constraints. {To this end, we extend the GCBF+ framework from \cite{zhang2024gcbf+} to incorporate robot connectivity in addition to safety. The GCBF+ framework uses a graph neural network (GNN) to learn a distributed control policy for collision avoidance, where edge features encode the essential information for maintaining safety (i.e., relative state information). In this work, we add edge features with the required connectivity information and modify the definition of the safe set to include connectivity requirements.} 

This work proposes a hierarchical control architecture in which a high-level planner can intervene and provide a mechanism to resolve deadlocks. Our proposed planner takes the environment information available to the MRS so far and proposes a high-level command in terms of a leader assignment for the MRS and a set of waypoints for it to navigate safely in obstacle environments. 
Motivated by the generalizability and low data requirements of pre-trained foundation models \cite{llms-zero-shot-reasoners} as well as the recent success of foundation models in assisting a control framework for complex robotics problems \cite{ren2024explore,huang2024grounded}, we explore the possibility of using text-based large language models (LLMs) and text-and-image-based vision language models (VLMs) as high-level planners to resolve deadlocks in MRS. 
Instead of \textit{proactively} using these models, whether directly for planning, translation, or reward design, we are instead interested in using foundation models \emph{reactively} to resolve a class of failure modes in low-level controllers for MRS, namely deadlocks. This also helps to ensure that the VLM or LLM does not lead to a violation of safety as it is taken care of by the provably safe low-level control policy. When a deadlock is detected, a VLM or an LLM is prompted to decide on a leader for the MRS and a set of waypoints for navigation to resolve it, conditioned on a top-view image-based description (for VLM) or text-based description (for LLM) of the so far observed environment. The MRS then reconfigures into a leader-follower formation to evade the deadlock situation. 
This temporary high-level assignment aims to move the MRS out of the deadlock so that the low-level controller can continue progressing toward the goal.

\textbf{Contributions}~ The contributions of the paper are as follows. 1) We propose a novel hierarchical control architecture to resolve deadlocks in an obstacle-cluttered environment for MRS (see Figure \ref{fig:algo overview}). For large-scale MRS, we propose a hierarchy of leaders for efficient deadlock evasion. 2) We compare the performance of various text-based LLMs and text-and-image-based VLMs in efficient deadlock resolution for MRS. We perform extensive experiments on a variety of MRS environments varying the number of agents from $5$ to $50$ with various foundation models. Our results demonstrate that both VLM and LLM-based high-level planners are effective at resolving deadlocks in MRS and more efficient {in terms of MRS goal-reaching rate in a given time budget} as compared to {grid-based planners such as A*}. Based on our observations, we provide a detailed discussion and possible future directions to improve the performance of foundation models as high-level planners for assisting low-level controllers in complex MRS problems.

\section{Related work}
Pre-trained LLMs have been shown to exhibit generalization to novel tasks without requiring updates to the underlying model parameters \cite{llms-few-shot-learners,llms-zero-shot-reasoners}. 
While originally intended for language tasks, pre-trained LLMs have since been adopted for use in robotics for planning and control design. For planning, a class of approaches has investigated the direct use of LLMs as planners by prompting them to generate sequences of actions by which a robot could accomplish a given task \cite{llms-zero-shot-planners, saycan, text2motion, inner-monologue, scalable-multi-robot-llms}; some methods rank the possible next action according to the probability of the LLM generating that action combined with the likelihood that the action will succeed \cite{saycan}, even iterating between LLM action proposals and estimates of individual action success probabilities to handle long-horizon tasks \cite{text2motion}. Instead, another class of methods relies on LLMs to translate from a natural language task description to a formal representation that can be provided as input to existing planners \cite{autotamp,nl2tl,translating-NL-to-PDDL-goals,pddl-planning-with-llms,llm+p}. More recently, VLMs have become popular in robotic applications due to their strong semantic reasoning capabilities \cite{ren2024explore}. VLMs have shown promising results in reasoning about future actions of robotic systems with partial environment information \cite{kwon2023toward,ma2022sqa3d,wen2023road,chen2024spatialvlm,gao2023physically,jiang2022vima}. In particular, VLMs have been successfully employed in robot navigation tasks \cite{shah2023navigation,dorbala2024embodied,shah2023lm}, even when long-horizon reasoning is required \cite{huang2024grounded}. 


\section{Problem formulation}
In this work, we design a distributed control framework for large-scale multi-robot systems (MRS) with multiple objectives. The MRS consists of $N$ robots navigating in an obstacle-cluttered environment to reach their goal locations $\{\pg_i\}_{i=1}^N$. The environment $\mathcal X\subset \mathbb R^2$ consists of stationary obstacles $\mathcal O_l\subset \mathbb R^2$ for $l\in \{1, 2, \dots, M\}$, which denote walls, blockades, and other obstacles in the path of moving agents. Each agent has a safety distance $r>0$ and a limited sensing radius $R > r$ such that the agents can only sense other agents or obstacles if they lie within their sensing radius. 
The agents use LiDAR to sense the obstacles, and {the observation data for each agent $i$ consists of $\nrays$ evenly-spaced LiDAR rays $y^{(i)}_j$ originating from each robot and measures the relative location of obstacles (see Appendix \ref{app: detailed problem} for more details).} 
The time-varying connectivity graph $\mathcal G(t) = (\mathcal V(t), \mathcal E(t))$ dictates the network among the agents and obstacles. Here, $\mathcal V(t) = \mathcal V^a\cup \mathcal V^o(t)$ denotes the set of nodes, where $\mathcal V^a = \{1, 2, \dots, N\}$ denotes the set of agents, $\mathcal V^o(t)$ is the collection of all LiDAR hitting points at time $t\geq 0$, and $\mathcal E(t)\subset\mathcal V^a\times\mathcal V$ denotes the set of edges, where $(i, j)\in\mathcal E(t)$ means the flow of information from node $j$ to agent $i$. In addition to safety, the resulting underlying graph topology for the MRS is required to remain connected so that robots can build team knowledge and share information, {where given a communication radius $R>0$, two agents $i,j$ are connected if $\|p_i-p_j\|\leq R$} (see Appendix \ref{app: detailed problem} for more details). The formal problem statement studied in this paper is described in the following. 


\begin{Problem}\label{prob: MRS safe connect perform}
Consider the multi-agent system 
with connected initial topology $\mathcal G(0)$, safety parameters $r > 0$, a sensing radius $R > 0$, a set of stationary obstacles $\{\mathcal O_j\}_{j=1}^M$, goal locations $\{\pg_i\}_{i=1}^N$, { and a terminal time $T_F > 0$}. 
Design a distributed control architecture such that
\begin{itemize}
    \item \textbf{Safety}: The agent maintains a safe distance from other agents and obstacles at all times, i.e., $\|p_i(t)-p_j(t)\| \geq 2r, \forall j\ne i$ and $\|y^{(i)}_j(t)\| > r,\;  \forall j = 1, \dots, \nrays$ for all $t\geq 0$;
    \item \textbf{Connectivity}: The graph $\mathcal G(t)$ remains connected at all times; 
    \item \textbf{Performance}: The agents reach their respective goals, i.e., $\lim\limits_{t \to T_F}\|p_i(t) - \pg_i\|= 0$. 
\end{itemize}
\end{Problem}

\section{Hierarchical control architecture}\label{sec: hierar control}
In an MRS problem with connectivity requirements, the presence of obstacles can lead to deadlock situations for the entire MRS, as illustrated in Figure \ref{fig:algo overview}. In this work, we propose a hierarchical control architecture consisting of a low-level control policy that accounts for safety and connectivity constraints and a high-level planner that assists the low-level controller with the goal-reaching requirement upon detection of a deadlock. We first describe how we detect the deadlocks. 

\textbf{Deadlock detection}~ In the proposed hierarchical architecture, the high-level planner is triggered upon detection of a deadlock, which we define as a situation when the average speed of the MRS falls below a minimum threshold $\delta_v>0$, i.e., $\frac{\sum_i\|\dot p_i\|}{N} < \delta_v$ and the average distance of the agents from their goals is at least $\delta_d>0$, i.e., $\frac{\sum_i\|p_i-\pg_i\|}{N} > \delta_d$. These criteria imply that the MRS is stuck in a deadlock due to the obstacles as the agents are not near their goals. Since the graph topology is connected, the average MRS speed can be computed through consensus updates. 

\textbf{Leader-follower formation}~When a deadlock is detected, the high-level planner assigns a leader among the $N$ agents along with a set of intermediate waypoints for the leader so that the leader does not get stuck in a deadlock due to obstacles on its path to its goal. 
We provide the details of the VLM/LLM-based high-level planner in Section \ref{sec: VLM planner}. Once a leader and its waypoints are obtained, the MRS reconfigures into a leader-follower formation. The leader-follower assignment is done by sequentially assigning the closest unassigned follower to its closest assigned agent as its leader. {The MRS remains in the leader-follower mode for a fixed time $T_{LF}>0$, which is a user-defined hyper-parameter.}
The complete leader-follower assignment algorithm is described in Appendix \ref{app: leader assign}.

\textbf{Multi-leader assignment using k-means clustering}~ In the cases of large-scale MRS, e.g., $N\geq 10$, assigning one leader to the MRS might lead to a sub-optimal performance. To this end, we decompose the MRS into sub-teams and assign a sub-leader to each of the sub-teams along with a  \textit{main} leader for the complete MRS. The decomposition of the MRS agents into $K\geq 1$ disjoint clusters $\mathcal V_k^a, k=1,...,K$ such that $\mathcal V^a = \cup_{k = 1}^K \mathcal V_k^a$ and $\mathcal V^a_i\cap \mathcal V^a_j= \emptyset$ for $i\neq j$, is performed based on the inter-agent distances using k-means clustering \cite{macqueen1967some}. The main leader $l_M\in \mathcal V^a$ is chosen as the agent with minimum distance to its goal. Once the clusters of agents $\{\mathcal V_k^a\}$ are obtained, a sub-leader $l_k$ is chosen based on the vicinity of the agents in $\mathcal V_{k}^a$ to the cluster of the main leader $\mathcal V_M^a$ (see Appendix \ref{app: leader assign} for more details). Note that for large-scale MRS, the high-level planner is utilized only for the waypoint assignment as the leader is assigned heuristically based on the distance to the goal locations. 

\subsection{Foundation model based high-level planner}\label{sec: VLM planner}
To initiate the leader-follower assignment process as explained in the previous section, it is necessary to designate a leader for the MRS. 
Based on the success of foundation models in a variety of robotic tasks that require spatial understanding, we explore their utility as the high-level planner for the leader and waypoint assignment. 

To use a pre-trained foundation model for leader and waypoint assignment, we provide the model with task-relevant context that is expected to be helpful when generating a decision. In particular, the prompt to the model consists of three main components: (1) the \textit{Task description}, (2) an \textit{Environment state}, and (3) the \textit{Desired output}. Next, we explain each of the prompt components in more detail. 
The exact prompts used in the experiments are provided in Appendix \ref{app: LLM actual prompts}.

\textbf{Task description}~ The initial part of the prompt consists of a description of the deadlock resolution problem for a multi-robot system. This includes the system requirements of maintaining safety, connectivity, and each agent reaching its assigned goal. Further, we include a description of the planner's role in providing high-level commands when the MRS is stuck in a deadlock. The description also includes the number of waypoints $P>0$ that the planner is supposed to suggest. This component of the prompt is created offline and is fixed for all calls. 

\textbf{Environment state}~ The environment state of the MRS is a necessary context to make a good leader assignment decision, so we encode it in a textual description that is included as a component of the prompt. Since the obstacle information depends on the roll-out of the system, we construct this part of the prompt online after a deadlock has been detected. For VLMs, at any given time instant $t_q$ when the VLM is queried, the environment state is constructed via a \texttt{base64} encoded \texttt{JPEG} image that includes the location of the agents, their goals, and the obstacles seen by the MRS so far for all $t\leq t_q$ whose information comes from the LiDAR data (see Section \ref{sec: gcbf low level} for more details). An example input image to the VLM is given in Figure \ref{fig:algo overview}. To assist the VLM with the precise locations of the agent and the goals, we provide their text description. 
For LLMs, the environment state is represented in text as the tuple: (Number of agents, Safety radius, Connectivity radius, Agent locations, Agent goals, Locations of observed obstacles), where the agent locations are $\{p_i(t_q)\}$, the goals $\{\pg_i\}$, and the observed obstacles $\{o_i(t)\}_{t\leq t_q}$.

\textbf{Desired output}~ Finally, we describe the desired output, both in terms of content and format. The high-level planner is responsible for choosing a leader and a set of waypoints for the leader. To help constrain the model's output and enable consistent output parsing, we request the generated response to be formatted as a \texttt{JSON} object with fields ``Leader'' and ``Waypoints'', with an expected output of the form $\{\text{``Leader''}: \text{Id}, \text{``Waypoints''}: [[x_1, y_1], \dots, [x_P, y_P]]\}$. 

\subsection{Distributed low-level policy}\label{sec: gcbf low level}
The high-level planner provides the leader and the waypoint information to the low-level controller. One desirable property of the low-level controller is scalability and generalizability to new environments (i.e., changing the number of agents and obstacles) while keeping the MRS safe and connected. Given the leader and goal information in terms of the immediate waypoint to follow, the low-level controller synthesizes an input $u_i$ to maintain connectivity, keep the system collision-free, and drive the system trajectories toward its goals (and in the case of the leader robot, toward its waypoint). While any low-level controller that can satisfy the requirements from Problem \ref{prob: MRS safe connect perform} can be used, we extend the distributed graph-based policy from \cite{zhang2024gcbf+}. Since the main focus of this paper is on the high-level planner, the details of the low-level controller are provided in Appendix \ref{app: low level policy}. We also show that the proposed hierarchical architecture does not get stuck in any deadlocks in the Appendix \ref{app: proof of complete}. 

\begin{figure}
    \centering
    \includegraphics[width=.4\columnwidth,trim={0 0 0.945in 0.945in}]{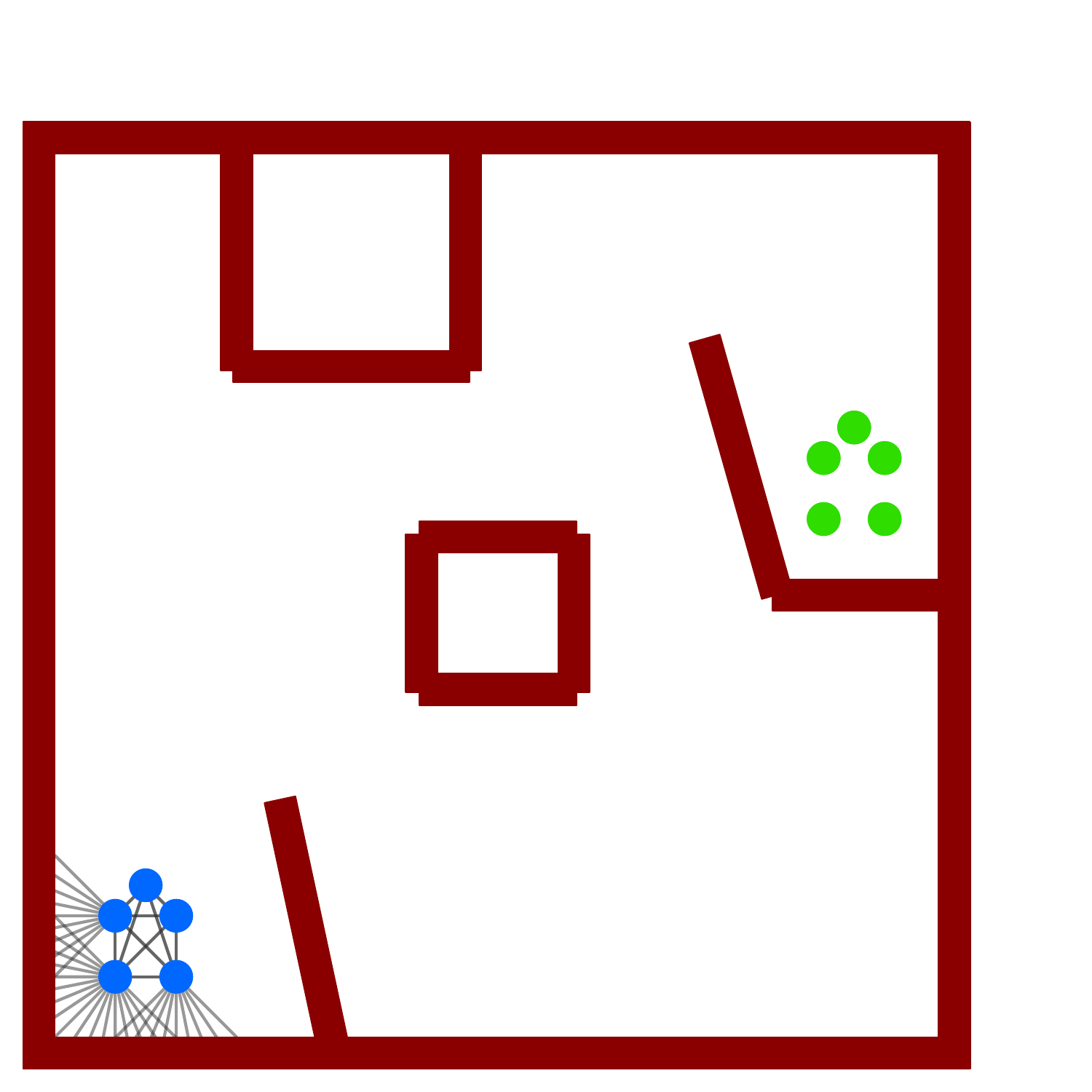}
    \hspace{10pt}
    \includegraphics[width=.4\columnwidth]{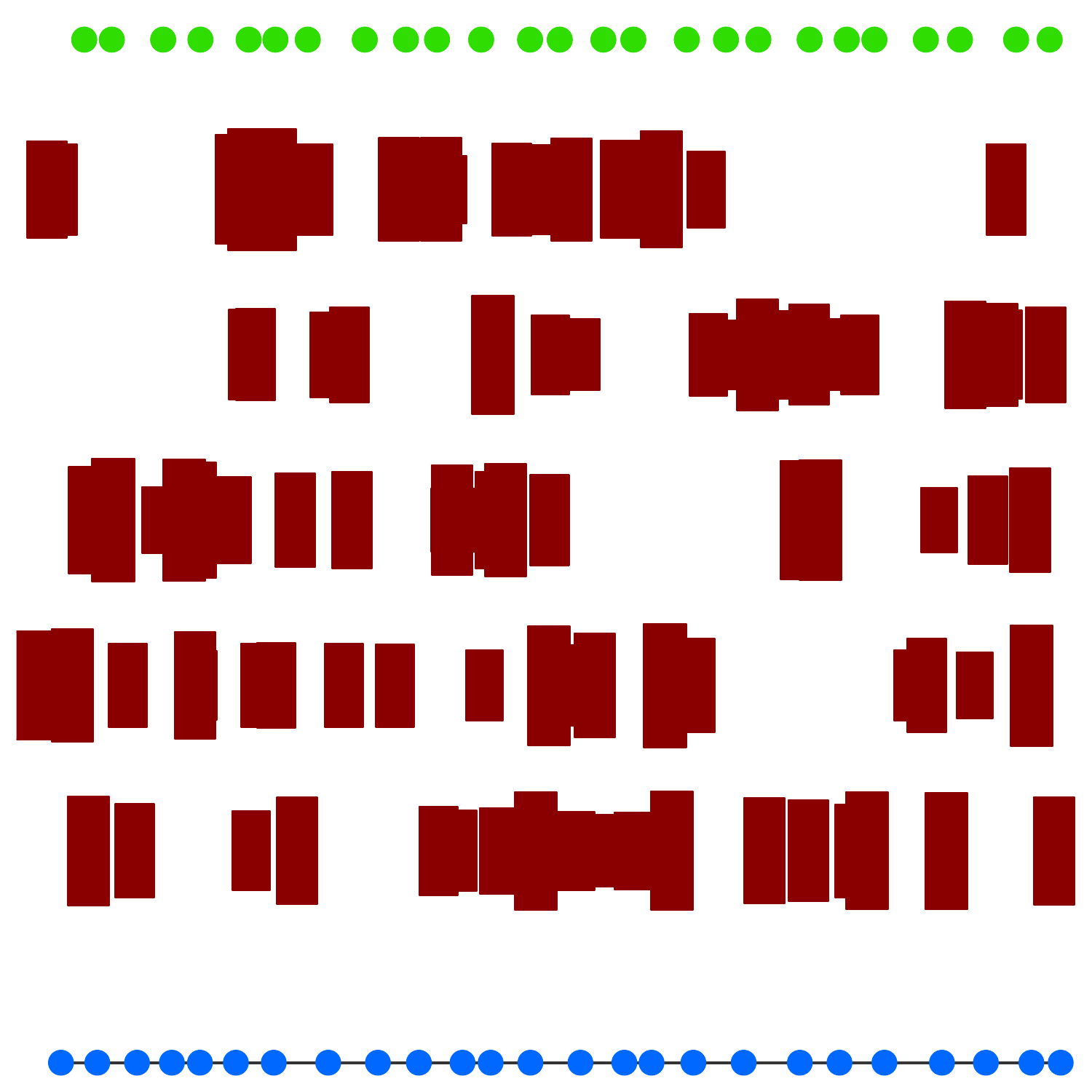}
    \caption{``Room'' environment with $N=5$ (left) and ``Maze'' environment with $N=25$ (right). The agents are shown in blue, the goals in green, and the obstacles in red color. }
    \label{fig:room env illust}
\end{figure}

\section{Evaluations}

To assess the effectiveness of the high-level planner, we perform extensive experiments on a variety of MRS environments with the objective of measuring the performance in terms of i) the reach rate or the percentage of agents reaching their goals; ii) the number of high-level interventions needed during the rollout; iii) the time taken in calling the high-level planner and iv) the tokens used in each intervention (for assessing the cost-efficiency of using proprietary pre-trained foundation models).

\subsection{Experiment setup}
We perform experiments on two sets of environments, namely structured hand-crafted environments (termed ``Room'') with a small number of agents, i.e., $N=5$, and unstructured maze-like environments (termed ``Maze'') with a large number of agents, i.e., $N=25$ or $50$ (see Figure \ref{fig:room env illust}). 

\textbf{``Room'' environments}: The ``Room'' represents an enclosed warehouse or an apartment scenario where the agents are required to reach another part of the environment while remaining inside the boundary and avoiding collisions with walls and other obstacles in the room. The agents start in one corner of the room and propagate their way through the obstacle environment to reach their respective destinations. The obstacles in the room environment are designed in such a manner that the low-level GCBF+ control policy invariably gets stuck in a deadlock, making it essential to use a high-level planner. We generate $20$ environments with random angles, lengths, and locations of the walls. 

\textbf{``Maze'' environment}~ The ``Maze'' environment consists of $N$ initial and goal locations and $M$ rectangular obstacles of randomly generated sizes and locations. For testing, we use $20$ Maze environments with $N = 25$ and $M = 100$, and $11$ Maze environments with $N=50$ and $M=375$. 

\textbf{Foundation models and baselines} 
We use \texttt{Claude3-Sonnet} (or simply, \texttt{Claude3S})\footnote{https://www.anthropic.com/news/claude-3-family}, \texttt{Claude3-Opus} (or simply, \texttt{Claude3O}), \texttt{GPT-4o}\footnote{https://platform.openai.com/docs/models} and \texttt{GPT4-Turbo} (or simply, \texttt{GPT4}) models as VLMs for the high-level planner, and \texttt{GPT4}, \texttt{GPT3.5}, \texttt{Claude2}\footnote{https://www.anthropic.com/news/claude-2}, and \texttt{Claude3O}
as the LLMs for the high-level planner.
In all the generated environments, the low-level controller leads to a deadlock and as a result, the reach rate for just the low-level controller is exactly $0$. Hence, we do not include that as a baseline. {We use an A*-based high-level planner and a "Random" high-level planner where the leader and the waypoints are drawn randomly, as baselines for comparison.}

\begin{figure*}
    \centering
    \textbf{Results for ``Room'' environments with 5 agents}\par
    \includegraphics[width=0.95\linewidth,trim={0 0.8in 0 0.0in}]{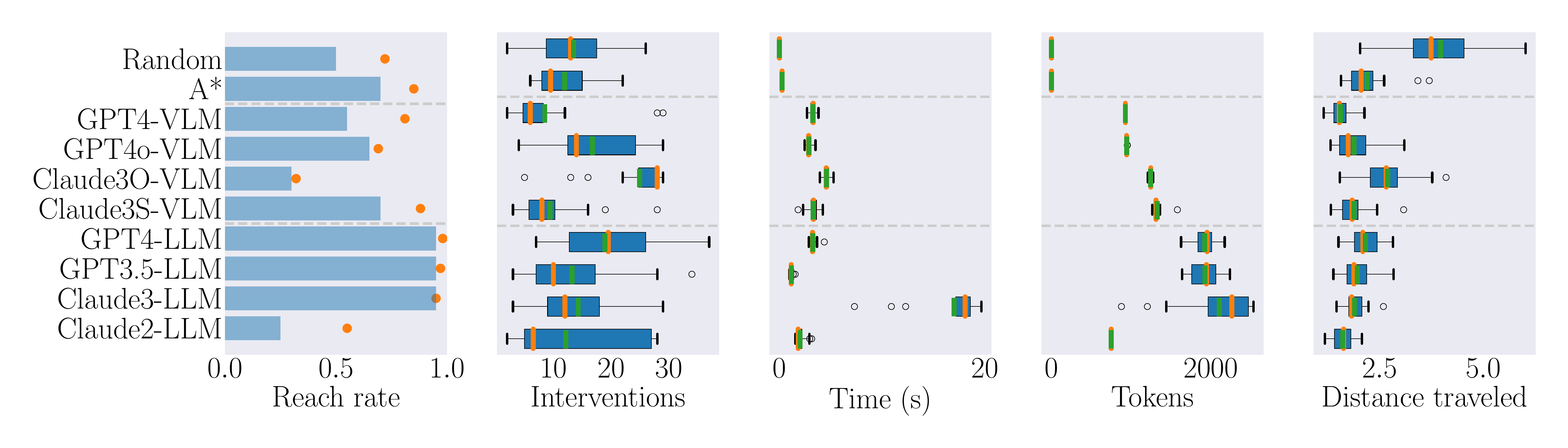}\par
     \textbf{Results for ``Maze'' environments with 25 agents}\par
    \includegraphics[width=0.95\linewidth,trim={0 0.8in 0 0.0in}]{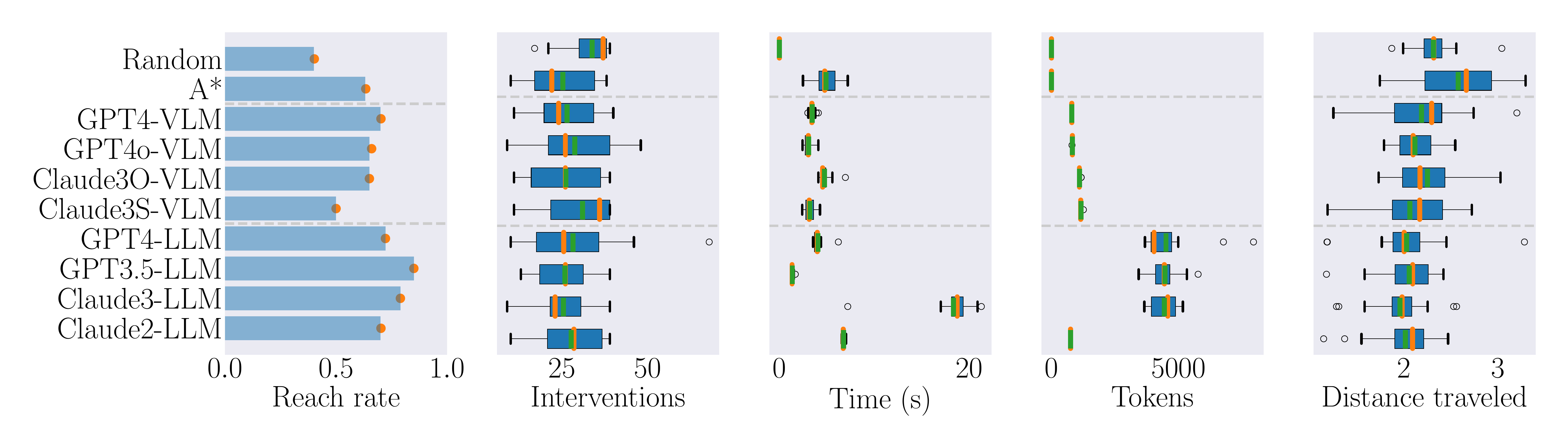}\par
     \textbf{Results for ``Maze'' environments with 50 agents}\par
    \includegraphics[width=0.95\linewidth,trim={0 0.8in 0 0.0in}]{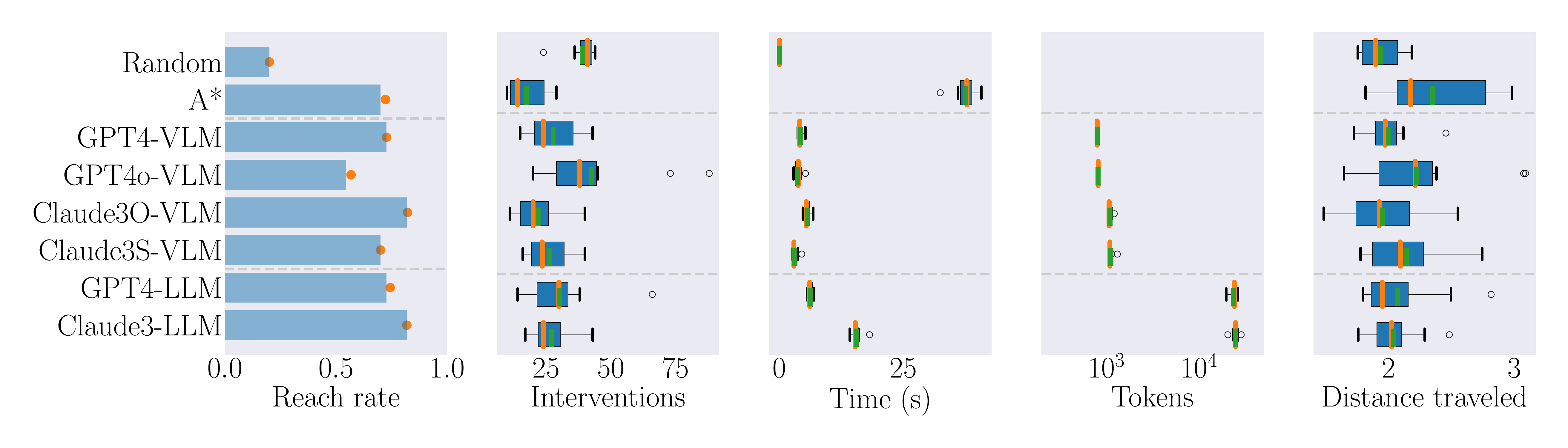}
    \caption{Performance of various high-level planners for ``Room'' environments with $N=5$ agents (Top plots), ``Maze'' environments with $N=25$ agents (Middle plots), and ``Maze'' environments with $N=50$ agents (Bottom plots). 
    From left to right: 1) The bar shows the ratio of the trajectories where \textbf{all} the agents reach their goals over the total number of trajectories, and the orange dot shows the ratio of agents that reach their goals over all agents; 2) Box plot of the number of times the high-level planner intervened; 3) Box plot of the time spent for each high-level planner intervention; and 4) Box plot for the input + output token per intervention. In the box plots, the median values are in orange and the mean values are in green.}
    \label{fig:result plots}
\end{figure*}

\textbf{Roll-out for evaluation} 
To evaluate the effectiveness of the high-level planner, we roll out MRS trajectories for a fixed number of steps $T = 4000$ {(i.e., $T_F = 1200s$)} for the Maze environment with $N=25$ and $T = 5000$ {(i.e., $T_F = 1500s$)} for the Maze environment with $N=50$, and $T=3000$ {(i.e., $T_F = 900s$)} for Room environments. As discussed in Section \ref{sec: hierar control}, the high-level planner intervenes when the MRS is stuck in a deadlock, defined according to the minimum average speed criteria. In this work, we use $\delta_v = 0.2$ and $\delta_d= 0.4$ as the threshold to define deadlocks. Once the high-level planner assigns a leader, the MRS remains in the same leader-follower configuration for { $T_{LF} = 100$ steps}. This is useful as it helps prevent Zeno behaviors, and sets an upper bound on the frequency at which the foundation model is queried, making the proposed framework applicable to real-time robotic applications.

\subsection{Results}\label{sec: results}
Figure \ref{fig:result plots} plots the various performance criteria for each of the test environments for the considered foundation models and baseline method. The broad observations and conclusions are summarized below. For the large-scale Maze environments with $N=50$ and $M=375$, the prompts for LLM-based planners \texttt{GPT3.5} and \texttt{Claude2} exceed the maximum allowed context window. Videos from the experiments and the code for running the experiments are included in the Supplementary material. 

\textbf{Foundation models are more effective than the grid-based and random method}: 
An important feature we note from the experiments is that foundation models do not require any in-context examples to achieve such high performance. This corroborates our motivation for using these models due to their low data requirement. When it comes to comparison to the A*-based planner, it is evident from all environments that foundation models achieve better performance in terms of higher average reach rate. For the small-scale Room environments, the mean reach rate for A* across 20 test environments is lower than LLMs such as \texttt{GPT4}, \texttt{GPT3.5} and \texttt{Claude3}, as well as \texttt{Claude3-Opus} LLM. For the large-scale Maze environments, {the mean reach rate for A* is lower than most of the foundation models. Note that the computation time of A* increases as the complexity of the environment in terms of its size and number of obstacles increases, while that of foundation models remains the same. Note also that a randomly chosen leader along with randomly chosen waypoints results in less than 0.5 reach rage in all cases, with its performance dropping as the number of agents becomes larger. This illustrates that the foundation models provide much better inference than a randomly generated one, resulting in a higher completion rate. } 

\textbf{LLMs outperform VLMs}:  
We observe that the text-based foundation models outperform text-and-image-based foundation models when it comes to performance in terms of reach rate. Moreover, from the number of high-level interventions, we infer that LLMs achieve similar or better performance than VLMs while requiring a lesser number of interventions. That signifies better leader and waypoint assignment by LLMs than VLMs, resulting in the completion of tasks with fewer interventions. 

\textbf{VLMs are cheaper and faster than LLMs for large-scale MRS}: It is evident that the average number of total tokens required for VLMs remains similar ($\approx 1000$) across the various environments. On the other hand, the tokens for LLMs increase with the increased complexity of the environment in terms of the number of obstacles. This directly affects the cost of using proprietary foundation models, as well as the speed of inference. This is also evident when we compare the time of intervention for the same model (e.g., \texttt{GPT4}) used as VLM and LLM. Furthermore, the variance in the average tokens used by VLMs across experiments is close to zero, while LLMs have a significant amount of variance. This is because as the MRS collects more information about its environment (i.e., observes newer obstacles), the text-based description becomes longer. On the other hand, all the new information can be plotted on the same-sized graphic, resulting in the same-sized input to the VLMs. This feature of VLMs is advantageous as it provides predictable costs, runtime, and scalability, especially for large-scale real-time systems. One way of reducing the prompt size for LLMs is to use a portion of the available information instead of the complete available information about the obstacles. We perform ablation on the available information {as well as on the utility of multi-leader assignment over just one leader in large-scale MRS} and report the results in Appendix \ref{app: ablation studies}.

\begin{figure}
    \centering
    \includegraphics[width=0.99\columnwidth,clip]{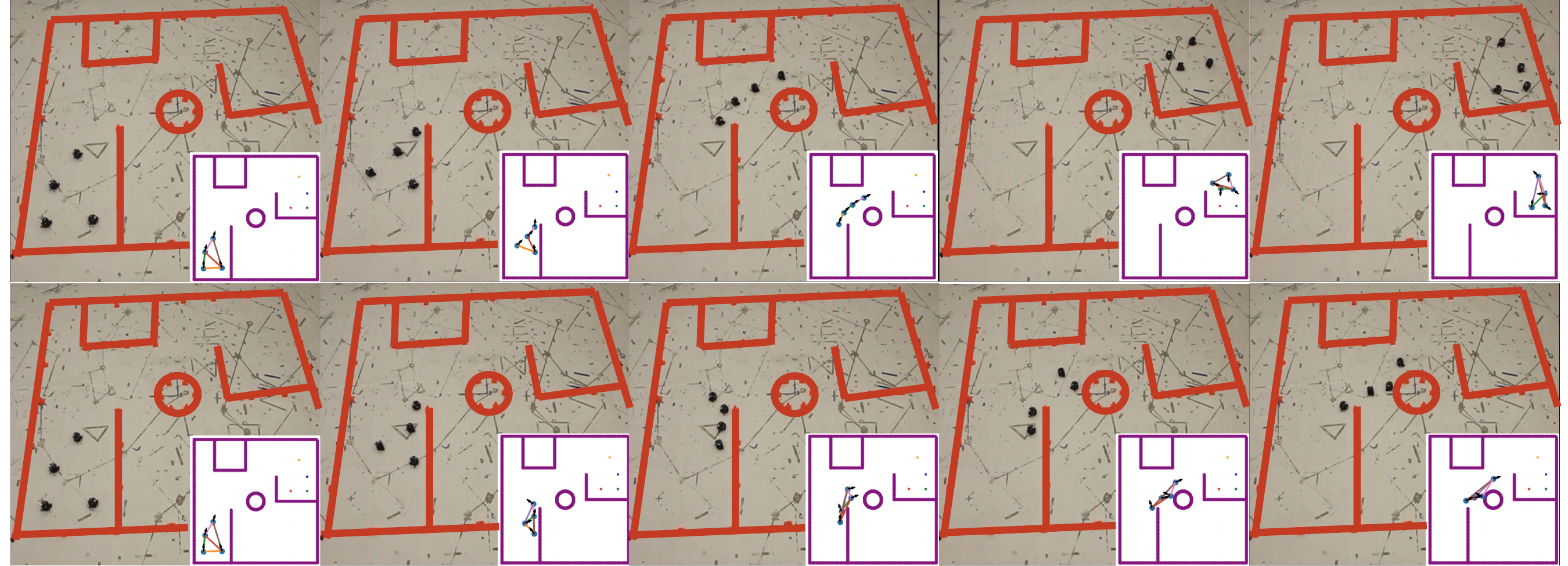}
    \caption{Snapshots from robot experiments with the proposed hierarchical control (top figures) and only the low-level controller (bottom figures). The hierarchical architecture leads to a successful run while just the low-level controller results in a deadlock.}
    \label{fig:exp snaps}
\end{figure}

\paragraph{Hardware demonstration}
We also validate our approach in hardware experiments on the first scenario, i.e., the room environment, using four Turtlebot3 ground robots. We use an off-board ground computer for sending control commands to the robots via ROS, while the state information of the robots is obtained using a Vicon motion capture system. Figure \ref{fig:exp snaps} shows the snapshots from the experiments under the proposed framework with \texttt{GPT3.5} LLM as the high-level planner as well as just the low-level controller without any high-level planner intervention. 
The experiments confirm the fact that only the low-level controller leads to deadlocks, while the proposed method is capable of completing the tasks. 



\section{Conclusions and Discussion}
In this work, we tested the hypothesis that text-based and text-and-image-based foundation models can be used as high-level planners for deadlock resolution in MRS with safety and connectivity constraints. We performed extensive experiments on a variety of foundation models to understand their utility in this task. In comparison to grid-based planners, the foundation models generally performed better, resulting in an affirmative answer to our hypothesis.
Our experiments also provided interesting observations and insights on the relative performance of LLMs and VLMs for small- and large-scale MRS problems, as well as their time and cost efficiency.  

The high performance of the zero-shot foundation model across various MRS environments is good evidence that they are promising high-level planners for deadlock resolution.  We are encouraged that in many cases LLM-based planners need to intervene less frequently, implying better intervention quality. However, the prompt size for LLMs increases significantly as the complexity of the environment increases. The prompt design used in this work is the result of manual trial and error while following generally accepted practices for prompt engineering. We note that the text-based environment description has much more flexibility as compared to the image-based description. One implication of this is the freedom to modify and optimize the prompt to obtain better performance. Another implication is the increased variability and, as a result, the high sensitivity of the performance of the model to the prompt design. 
Recently, there has been an increasing interest in automatic prompt optimization for black-box LLMs to find the best prompt design for a given task \cite{prompt-optimization-evolutionary,promptagent,promptbreeder} with results showing significant performance improvement over human-designed prompts. Future work includes applying such techniques to the problem of deadlock resolution to maximize the performance of LLMs as a high-level planner.

\section{Limitations}
One limitation of using a proprietary foundation model as a high-level planner is the time it takes to query the model and receive a response. For real-time intervention, fast performance can often be a prerequisite. 
Future work involves investigating 
possible methods to improve the runtime, such as using smaller, open-source models, running locally, fine-tuned to the specific task, or using embedding models that can capture semantic context with significantly lower runtime. In this work, the high-level planner intervenes according to a heuristic threshold of average agent velocity, which may not be an optimal indicator of the failure mode of the low-level controller. 
If the system can better anticipate deadlocks, the high-level planner can intervene earlier. Future work includes embedding the local environment information, either as text or image, to classify whether a deadlock is likely to occur.

\bibliographystyle{IEEEtran}
\bibliography{refs.bib} 

\newpage
\clearpage

\appendix

\section{Detailed problem formulation}\label{app: detailed problem}

We start with describing the dynamics of the individual robots (referred to as agents henceforth), and then, we list the individual as well as the team objective for the system. 
The agent dynamics are given by $\dot x_i = f(x_i) + g(x_i)u_i$, where $f, g$ are locally Lipschitz continuous functions with $x_i\in\mathcal X\subset \mathbb R^{n_{x}}$ denoting agents' state space and $u_i\in\mathcal U\subset \mathbb R^{n_{u}}$ the control constraint set for $i\in \{1, 2, \dots, N\}$.\footnote{In this work, we consider robots modeled using single integrator dynamics operating in 2D plane, i.e., $n_x=n_u=2$}. The state $x_i$ consists of the position $p_i\in \mathbb R^2$ in the global coordinates along with other states, such as the orientation and the velocity of the agent $i$. 
The state space $\mathcal X$ consists of stationary obstacles $\mathcal O_l\subset \mathbb R^2$ for $l\in \{1, 2, \dots, M\}$, denoting walls, blockades and other obstacles in the path of the moving agents. Each agent has a limited sensing radius $R > 0$ and the agents can only sense other agents or obstacles if it lies inside its sensing radius. 
The agents use LiDAR to sense the obstacles, and the observation data for each agent consists of $\nrays$ evenly-spaced LiDAR rays originating from each robot and measures the relative location of obstacles. We denote the $j$-th ray from agent $i$ by $y_j^{(i)}\in\mathcal X$ for $j\in \{1, 2, \dots, \nrays\}$ that carries the relative position information of the $j-$th LiDAR hitting point to agent $i$, and zero padding for the rest of the states. 

The time-varying connectivity graph $\mathcal G(t) = (\mathcal V(t), \mathcal E(t))$ dictates the network among the agents and obstacles. Here, $\mathcal V(t) = \mathcal V^a\cup \mathcal V^o(t)$ denotes the set of nodes, where $\mathcal V^a = \{1, 2, \dots, N\}$ denotes the set of agents, $\mathcal V^o(t)$ is the collection of all LiDAR hitting points at time $t\geq 0$, and $\mathcal E(t)\subset\mathcal V^a\times\mathcal V$ denotes the set of edges, where $(i, j)\in\mathcal E(t)$ means the flow of information from node $j$ to agent $i$. We denote the time-varying adjacency matrix for agents by $A(t)\in\mathbb R^{N\times N}$, where $A_{ij}(t) = 1$ if $(i, j)\in\mathcal E(t), i, j\in\mathcal V^a$, and $0$ otherwise. The set of \textit{all} neighbors for agent $i$ is denoted as $\mathcal N_i(t) \coloneqq \{j \; |\; (i, j)\in\mathcal E(t)\}$, while the set of \textit{agent} neighbors of agent $i$ is denoted as $\mathcal N_i^a(t) \coloneqq\{j\;|\; A_{ij}(t) = 1\}$. The MRS is said to be connected at time $t$ if there is a path between each pair of agents $(i, j), i, j\in\mathcal V^a$ at $t$. One method of checking the connectivity of the MRS is through the Laplacian matrix, defined as $L(A(t)) \coloneqq D(t) - A(t)$, where $D(t)$ is the degree matrix defined as $D_{ij}(t) = \sum\limits_{j}A_{ij}(t)$ when $i = j$ and $0$ otherwise. From \cite[Theorem 2.8]{mesbahi2010graph}, the MRS is connected at time $t$ if and only if the second smallest eigenvalue of the Laplacian matrix is positive, i.e., $\lambda_2(L(A(t))) > 0$. 

\section{Low-level control policy}\label{app: low level policy}

Any low-level controller that can satisfy the requirement from Problem \ref{prob: MRS safe connect perform} can be used as the low-level control policy, such as one from a distributed CBF-QP method \cite{wang2017safety} or a learned control policy. Since the low-level control policy is supposed to be distributed with low computational complexity, we choose to use the recently proposed learning-based GCBF+ controller from \cite{zhang2024gcbf+}. Here, we briefly review the GCBF+ controller and present the details of how the safety and connectivity constraints are encoded. 

\subsection{Graph control barrier functions (GCBF)}

Given sensing radius $R$ and safety distance $r$, define $N_s - 1\in\mathbb N$ as the maximum number of neighbors that each agent can have while all the agents in the neighborhood remain safe. Define $\tilde{\mathcal N}_i$ as the set of $N_s$ closest neighboring nodes to agent $i$ which also includes agent $i$ and $\bar{x}_{\tilde{\mathcal N}_i}$ as the concatenated vector of $x_i$ and the neighbor node states with fixed size $N_s$ that is padded with a constant vector if $|\tilde{\mathcal N_i}|<N_s$. Considering only collision avoidance constraints, the safe set $\mathcal S_N\in\mathcal X^N$ can be defined as:
\begin{align*}
    \mathcal S_N\coloneqq\Big\{ \bar{x} \in \mathcal{X}^N \;\Big|&\; \Big( \norm{y_j^{(i)}} > r,\; \forall i\in \mathcal V^a,\forall j\in n_\mathrm{rays} \Big) \nonumber\\
    & \bigwedge \Big( \min_{i,j \in\mathcal V^a, i \neq j} \norm{p_i - p_j} > 2r \Big) \Big\},
\end{align*}
where $\bar x$ denotes the joint state vector for the MRS.
The unsafe, or avoid set can be defined accordingly as $\mathcal A_N=\mathcal X^N\setminus\mathcal S_N$. 
We now introduce the notion of GCBF for encoding safety for MAS. 
Considering the smoothness of GCBF, 
we impose the condition that for a given agent $i \in V_a$, a node $j$ where $\norm{ p_i - p_j } \geq R$ does not affect the GCBF $h$. Specifically, for any neighborhood set $\mathcal{N}_i$, let $\mathcal{N}^{<R}_i$ denote the set of neighbors in $\mathcal{N}_i$ that are strictly inside the sensing radius $R$ as
\begin{equation}
    \mathcal{N}^{<R}_i \coloneqq \{ j : \norm{p_i - p_j} < R,\; j \in \mathcal{N}_i \}.
\end{equation}
Using these notations, the notion of GCBF is defined in \cite{zhang2024gcbf+} as:

\begin{Definition}[\textbf{GCBF}]\label{def: gcbf}
A continuously differentiable function $h : \mathcal{X}^M \to \mathbb{R}$ is termed as a Graph CBF (GCBF) if there exists an extended class-$\mathcal K$ function $\alpha$ and a control policy $\pi_i: \mathcal{X}^M \to \mathcal{U}$ for each agent $i \in V_a$ of the MAS such that, for all $\bar{x} \in \mathcal{X}^N$ with $N \geq M$,
\begin{equation}\label{eq:graph CBF}
        \dot h(\bar{x}_{\mathcal{N}_i}) +\alpha( h( \bar{x}_{\mathcal{N}_i} ) )\geq 0, \quad \forall i \in V_a
\end{equation}
where
\begin{equation} \label{eq:hdot_def}
    \dot h(\bar x_{\mathcal N_i}) = \sum_{j\in 
    \mathcal{N}_i}\frac{\partial h(\bar x_{\mathcal N_i})}{\partial x_j}\left(f(x_j) + g(x_j) u_j \right),
\end{equation}
for $u_j = \pi_j(\bar x_{\mathcal N_j})$, and the following two conditions hold:
\begin{itemize}
    \item The gradient of $h$ with respect to nodes $R$ away is $0$, i.e.,
    \begin{equation} \label{eq:ass_deriv_zero}
        \frac{\partial h}{\partial x_j}( \bar{x}_{\mathcal{N}_i} ) = 0, \quad \forall j \in \mathcal{N}_i \setminus \mathcal{N}^{<R}_i.
    \end{equation}
    \item The value of $h$ does not change when restricting to neighbors that are in $\mathcal{N}^{<R}_i$, i.e.,
    \begin{equation} \label{eq:ass_same_h}
        h( \bar{x}_{\mathcal{N}_i} ) = h( \bar{x}_{\mathcal{N}^{<R}_i} ).
    \end{equation}
\end{itemize}
\end{Definition}

It is proved in \cite[Theorem 1]{zhang2024gcbf+} GCBF certifies the forward invariance of its $0$-superlevel set under control inputs in the set
\begin{equation}
    \mathcal{U}^N_{\mathrm{safe}} \coloneqq \left\{ \bar{u} \in \mathcal{U}^N\; \Big|\; \dot h( \bar{x}_{\mathcal{N}_i} ) + \alpha(h(\bar{x}_{\mathcal{N}_i})) \geq 0,\, \forall i \in V_a \right\}.
\end{equation}
We start with this formulation and modify it to additionally account for the connectivity requirement, as explained below. 

\subsection{Learning GCBF with safety and connectivity constraints}

Following \cite{zhang2024gcbf+}, we use graph neural networks (GNN) to parameterize GCBF. For agent $i$, the input features of the GNN contain the node features $v_i$ and $v_j$ for $j\in\mathcal N_i$, and edge features $e_{ij}$ for $j\in\mathcal N_i$. The node features $v_i\in \mathbb R^{\rho_v}$ encode information specific to each node. In this work, we take $\rho_v = 3$ and use the node features $v_i$ to one-hot encode the type of the node as either an agent node, goal node or LiDAR ray hitting point node. The edge features $e_{ij}\in \mathbb R^{\rho_e}$, where $\rho_e > 0$ is the edge dimension, are defined as the information shared from node $j$ to agent $i$, which depends on the states of the nodes $i$ and $j$. 
Since the safety objective depends on the relative positions, one component of the edge features is the relative position $p_{ij} = p_j - p_i$. The rest of the edge features can be chosen depending on the underlying system dynamics, e.g., relative velocities for double integrator dynamics. However, apart from the safety constraints considered in \cite{zhang2024gcbf+}, we also consider the connectivity constraints. Therefore, the design of the node features and edge features needs to be modified for adding the connectivity information. To this end, given the desired connectivity of the MRS in terms of the \textit{desired} adjacency matrix $A^d$ where the desired adjacency matrix is designed such that the MRS is connected, we add the connectivity information in the edge features of GCBF. In particular, we append the edge features with $[0, 1]^\top$ in $e_{ij}$ if the agents $(i,j)$ are required to be connected, i.e., $A^d_{ij} = 1$, and $[1,0]^\top$ if they are not required to be connected, i.e., $A^d_{ij}= 0$. Furthermore, we add the connectivity constraint in the GCBF by redefining the safe and the unsafe sets corresponding to the required connectivity, such that the safe set is defined as
\begin{align}\label{eq: safe and connect set S}
    \mathcal S_N^c\coloneqq\Big\{ & \bar{x} \in \mathcal{X}^N \;\Big|\; \Big( \norm{y_j^{(i)}} > r,\; \forall i\in \mathcal V_a,\forall j\in n_\mathrm{rays} \Big)\nonumber \\
    & \bigwedge \Big( \min_{i,j \in \mathcal V_a, i \neq j} \norm{p_i - p_j} > 2r \Big) \nonumber\\
    &\bigwedge\Big( \max_{i,j \in \mathcal V_a, A^d_{ij} = 1} \norm{p_i - p_j} < R \Big)\Big\}.
\end{align}
Consequently, the unsafe, or avoid set with the connectivity constraint is defined as $\mathcal {A}^c_{N} = \mathcal X^N\setminus\mathcal S^c_{N}$. 
Since the GCBF $h$ certifies the forward-invariance of its $0$-superlevel set, the safety and connectivity constraints are satisfied. 

The training framework is the same as \cite{zhang2024gcbf+}. In the original GCBF+ training framework, it is essential that the initial and goal locations are safe. In the current work, we also need to make sure that the initial conditions and the goal locations sampled for training the GCBF satisfy the MRS connectivity condition, in addition to the safety condition in the original GCBF+ framework. To this end, we sample the initial and goal locations such that their corresponding graph topology are connected, and define the desired adjacency matrix $A^d = A(0)$. The same loss function from \cite{zhang2024gcbf+} is used to train the distributed control policy, with the safe set definition modified as per \eqref{eq: safe and connect set S}.



\section{Leader-follower and temporary goal assignment}\label{app: leader assign}
\subsection{Leader-follower for small MRS}
The follower assignment is carried out in an iterative way. Let $\mathcal V_\mathrm{lead}(t, k)$ be the set of agents that have been assigned as a leader at time $t$, iteration $k$, initiated as $\mathcal V_\mathrm{lead}(t, 0) = \{i_{\mathrm{lead},0}\}$, where $i_{\mathrm{lead},0}\in \mathcal V$ is the leader agent. Then, the $k-$th follower with $k\geq 1$ is chosen as 
\begin{align}\label{eq:follow k}
    i_{\mathrm{follow},k} = \argmin_{j\in \mathcal V\setminus \mathcal V_\mathrm{lead}(t, k-1)}\min_{i\in \mathcal V_\mathrm{lead}(t, k-1)}\|p_i-p_j\|,
\end{align}
and this follower is added to the set of the leaders, i.e., $\mathcal V_\mathrm{lead}(t, k) = \mathcal V_\mathrm{lead}(t, k-1)\cup \{i_{\mathrm{follow},k}\}$. The leader for the $k-$th follower is given as $i_{\mathrm{lead},k} = \argmin\limits_{i\in \mathcal V_\mathrm{lead}(t, k-1)}\|p_{i_{\mathrm{follow},k}}-p_i\|$. The process is repeated till each agent $i$ is assigned an agent $i_\mathrm{lead}$ to follow. 
Next, for each agent $i$ that is a given minimum distance away from its goal, i.e., if $\|p_i-\pg_i\|\geq d_\mathrm{min}$ for some $d_{min}> 0$, their temporary goal is chosen as the location of their leaders, i.e., $\pgtemp_i = p_{i_\mathrm{lead}}$. 

\subsection{Leader-follower for large MRS}\label{sec:multi leader assign}
As discussed in Section \ref{sec: hierar control}, for large-scale MRS, i.e., $N\geq N_M = 10$, a main leader is assigned for the MRS, and sub-leaders are assigned for each of the clusters. The agents in clusters will undergo a leader-follower formation with their respective sub-leaders while these sub-leaders will follow the main leader. The complete leader-follower assignment algorithm is given in Algorithm \ref{alg: temp goal assign}. 

\SetKwComment{Comment}{/* }{ */}

\begin{algorithm}
\caption{Leader-follower and temporary goal assignment}\label{alg: temp goal assign}
\KwData{$\{p_i\}, K, d_{min}, N, N_M$}
\KwResult{$l_M, \{l_k\}, \{\pg_{\text{temp}}\}$}

\Comment{Find the main leader}

$l_M = \argmin\|p_i-\pg_i\|$

\Comment{Find the clusters using k-means clustering}

\eIf{$N< N_M$
}{
$\{\mathcal V_k^a\} = \{\mathcal V^a\}$
}{
$\{\mathcal V_k^a\}$ = kmeans($\{p_i\}, K$)
}

\Comment{Find sub-leaders for each of the cluster}
\For{k in $\textnormal{range}$(K)}{
$l_k = \argmin_{i\in \mathcal V^a_k}\min_{j\in \mathcal V_M^a}\|p_i-p_j\| $
}

\Comment{Assign temporary goals to each agent}

\For{$k \in [1, 2, \cdots, K]$}
{

\Comment{Initial set of leaders to be followed in cluster $\mathcal V^a_k$}

$\mathcal V_\mathrm{lead}(k) = \{l_k\}$ 

\Comment{Assign main leader's location as the temporary goal for cluster leader}

$\pg_{\mathrm{temp}, l_k} = p_{l_M}$ 

\For{$i\in \textnormal{range}|\mathcal V_a^k|$}{

\Comment{Find closest agent to the set of leader as the new follower}

$i_{\mathrm{follow}} = \argmin_{j\in \mathcal V_k^a\setminus \mathcal V_\mathrm{lead}(k)}\min_{l\in \mathcal V_\mathrm{lead}(k)}\|p_l-p_j\|$ 

\Comment{Find leader for this follower}

$k_{\text{lead}} = \text{arg}\min_{j\in \mathcal V_\mathrm{lead}(k)}\|p_{i_\text{follow}}-p_j\|$

\Comment{Assign temporary goal to the follower}

\eIf{$\|p_{i_\mathrm{follow}}-\pg_{i_\mathrm{follow}}\|\geq d_{min}$}{
$\pg_{i_\mathrm{follow}, \mathrm{temp}} = p_{k_\mathrm{lead}}$
}
{
$\pg_{i_\mathrm{follow},\mathrm{temp}} = \pg_{i_\mathrm{follow}}$
}

\Comment{Update the set of leaders}

$\mathcal V_\mathrm{lead}(k) =\mathcal V_\mathrm{lead}(k)\cup \{i_{\text{follow}}\}$

}
}

\end{algorithm}

\newpage
\clearpage

\section{VLM and LLM prompts}\label{app: LLM actual prompts}
\subsection{Task and output description prompts}
\begin{figure}[hb]
    \centering
    \includegraphics[width=1\columnwidth]{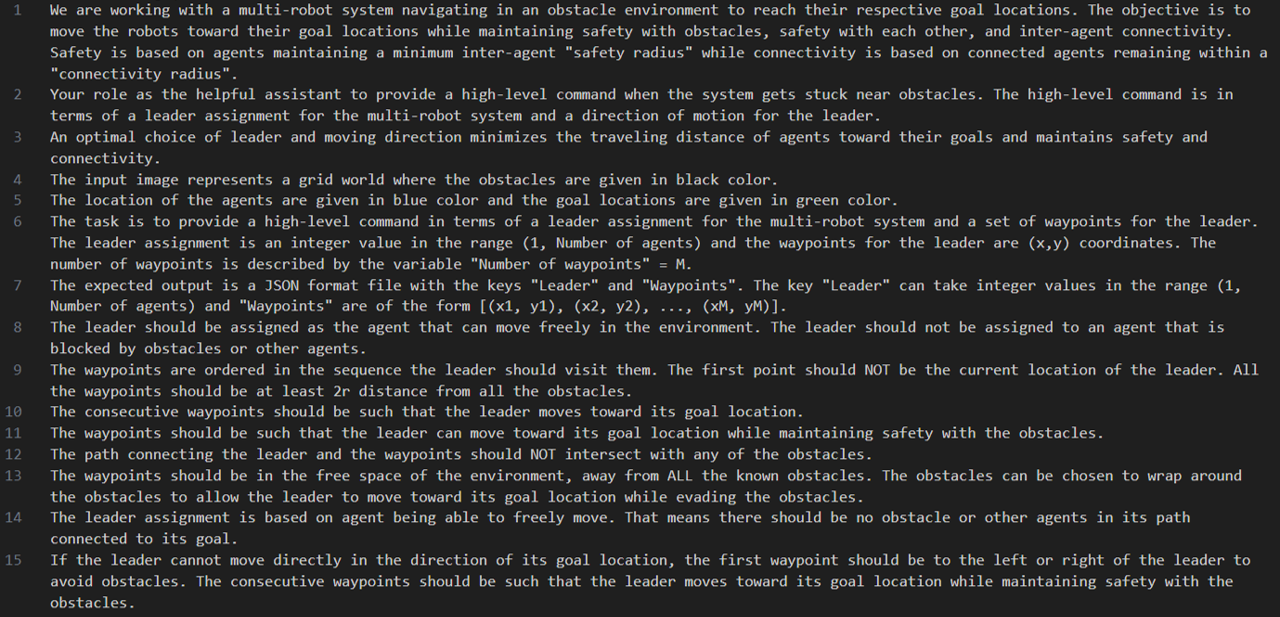}
    \caption{Description prompts used for vision-based (i.e., VLMs) high-level planners.}
    \label{fig:vlm prompts}
\end{figure}

\begin{figure}[hb]
    \centering
    \includegraphics[width=1\columnwidth]{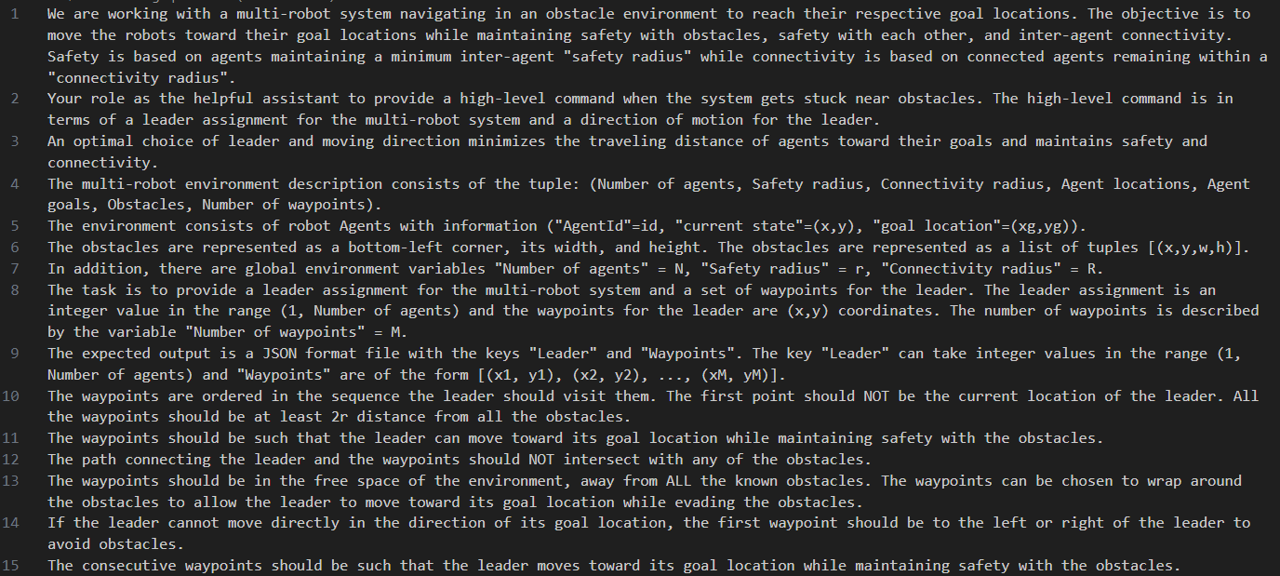}
    \caption{Description prompts used for text-based (i.e., LLMs) high-level planners.}
    \label{fig:llm prompts}
\end{figure}

The task description prompts used for VLMs are given in Figure \ref{fig:vlm prompts} while those used for LLMs are given in Figure \ref{fig:llm prompts}. 

\subsection{Environment description}
\begin{figure}[hb]
    \centering
    \includegraphics[width=0.7\columnwidth]{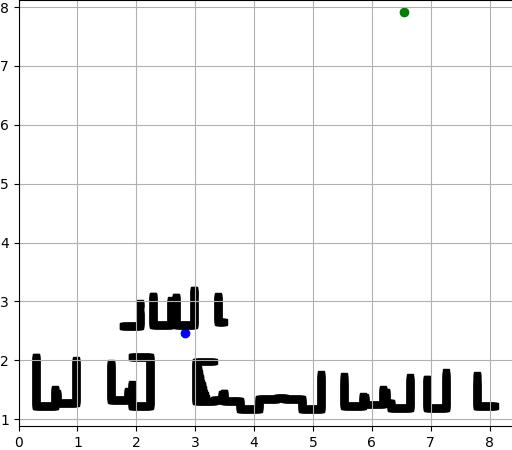}
    \caption{Environment prompt for VLM for "Maze" environment with $N=50$ and $M=375$.  }
    \label{fig:vlm env prompt}
\end{figure}

\begin{figure}[hb]
    \centering
    \includegraphics[width=1\columnwidth]{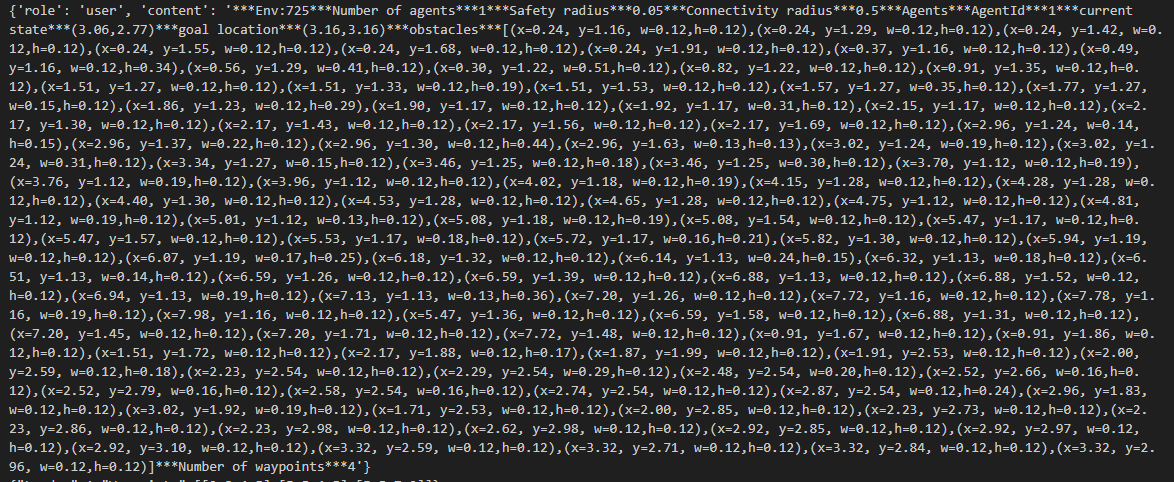}
    \caption{Environment prompt for LLM for "Maze" environment with $N=50$ and $M=375$. }
    \label{fig:llm env prompt}
\end{figure}

The environment description prompts used for VLMs are given in Figure \ref{fig:vlm env prompt} while those used for LLMs are given in Figure \ref{fig:llm env prompt}. For VLMs, an additional text prompt is appended at the end with the location of the agent(s) and goal(s) provided in the image prompt to aid the VLM with waypoint assignment. 

\subsection{Output of foundation models}
The output of the LLM or VLM is a JSON object. As an example, for the input in Figure \ref{fig:vlm env prompt}, the output from \texttt{Claude3-Opus} VLM is: $\{\textrm{``Leader''}:1,\textrm{``Waypoint''}:[[2.8,4.5],[5.5,4.5],[5.5,7.9]]\}$.

\newpage

\clearpage

{
\section{Additional experimental data}

In Figure \ref{fig: add result plots}, we report the mean distance traveled by the MRS under various high-level planners normalized by the mean initial distance of MRS from the goal locations. 
For ``Room" environment, ``random" baseline has the highest mean distance traveled among all methods, while having a relatively lower reach rate, compared to some of the foundation models. 
There does not seem to be much variation in the distance traveled among LLM-based high-level planners for ``Maze" environments. While GPT4-VLM has a relatively higher mean traveled distance, the A*-based baseline has the highest distance traveled for ``Maze'' environment with $N=25$. For ``Room'' environment, \texttt{Claude3O-VLM} has the highest mean distance traveled by MRS. An interesting observation here is that unlike other performance metric which seem to have a stronger correlation among themselves, the mean distance traveled does not seem to have such property. As an example, for ``Room'' environment and ``Maze'' environment with $N=50$ with VLMs as high-level planners, the distance traveled seems to follow the same pattern as the number of calls. However, the same cannot be said about ``Maze'' environment with $N=25$. 

\begin{figure*}[h]
    \centering
    \textbf{Results for ``Room'' environments with 5 agents}\par
    \includegraphics[width=0.9\linewidth,trim={0 0.8in 0 0.0in}]{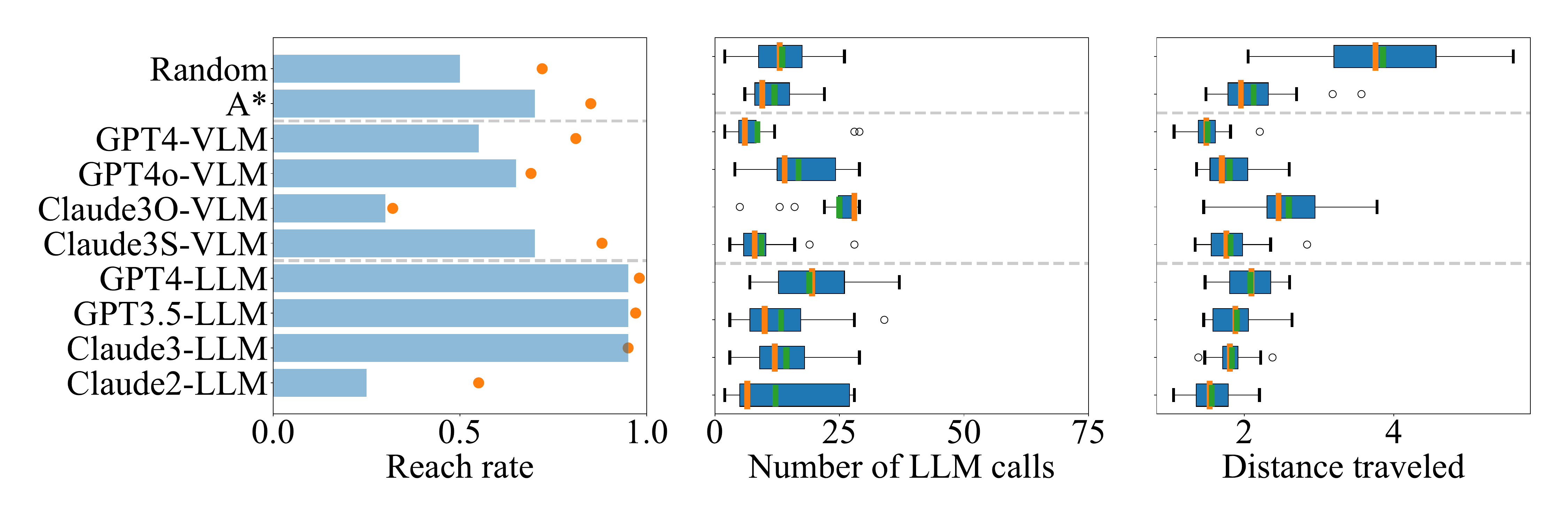}\par
     \textbf{Results for ``Maze'' environments with 25 agents}\par
    \includegraphics[width=0.9\linewidth,trim={0 0.8in 0 0.0in}]{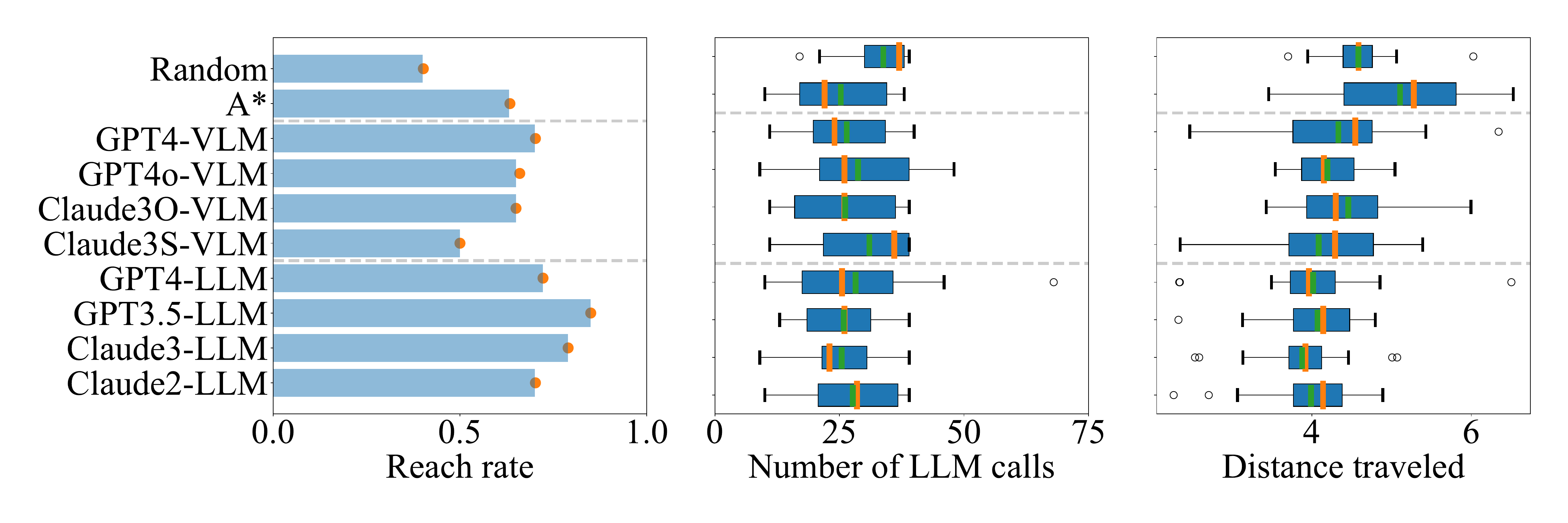}\par
     \textbf{Results for ``Maze'' environments with 50 agents}\par
    \includegraphics[width=0.9\linewidth,trim={0 0.8in 0 0.0in}]{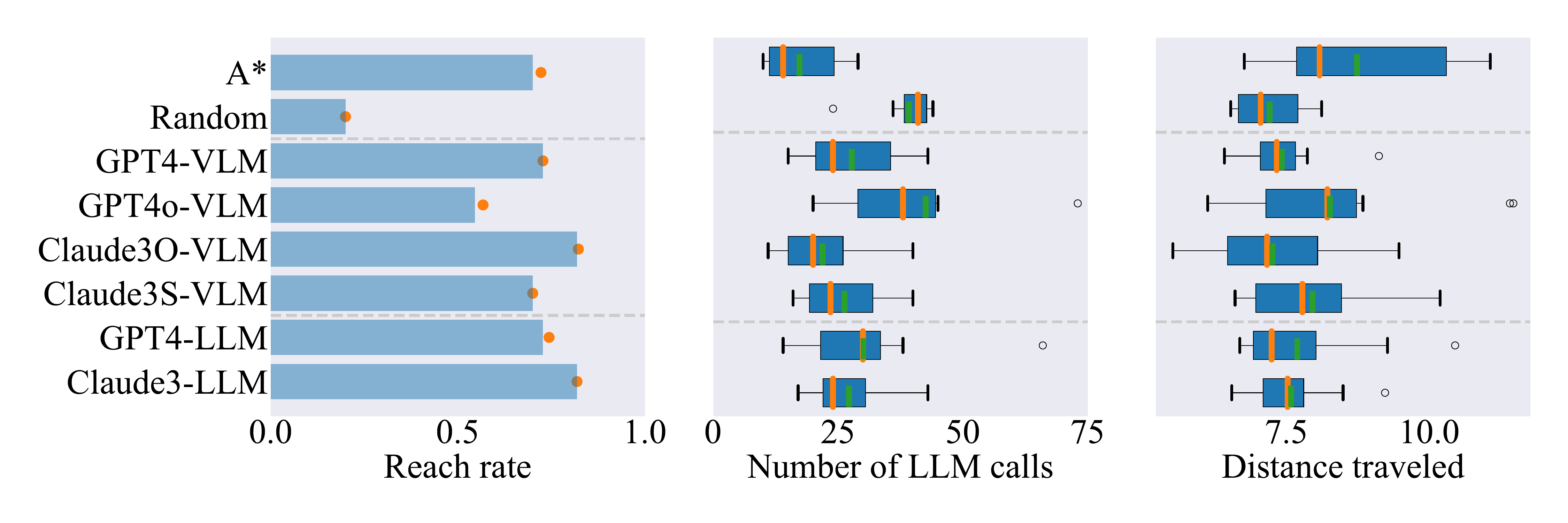}
    \caption{Performance of various high-level planners for ``Room'' environments with $N=5$ agents (Top plots), ``Maze'' environments with $N=25$ agents (Middle plots), and ``Maze'' environments with $N=50$ agents (Bottom plots). 
    From left to right: 1) The bar shows the ratio of the trajectories where \textbf{all} the agents reach their goals over the total number of trajectories, and the orange dot shows the ratio of agents that reach their goals over all agents; 2) Box plot of the number of times the high-level planner intervened; and 3) Box plot for the mean traveled distance by the multi-robot system normalized by the mean initial distance from the goal locations. In the box plots, the median values are in orange and the mean values are in green.}
    \label{fig: add result plots}
\end{figure*}

}

\section{Ablation studies}\label{app: ablation studies}

\subsection{Effect of partial environment infromation}

\begin{figure*}[h]
    \centering
    \includegraphics[width=0.9\linewidth]{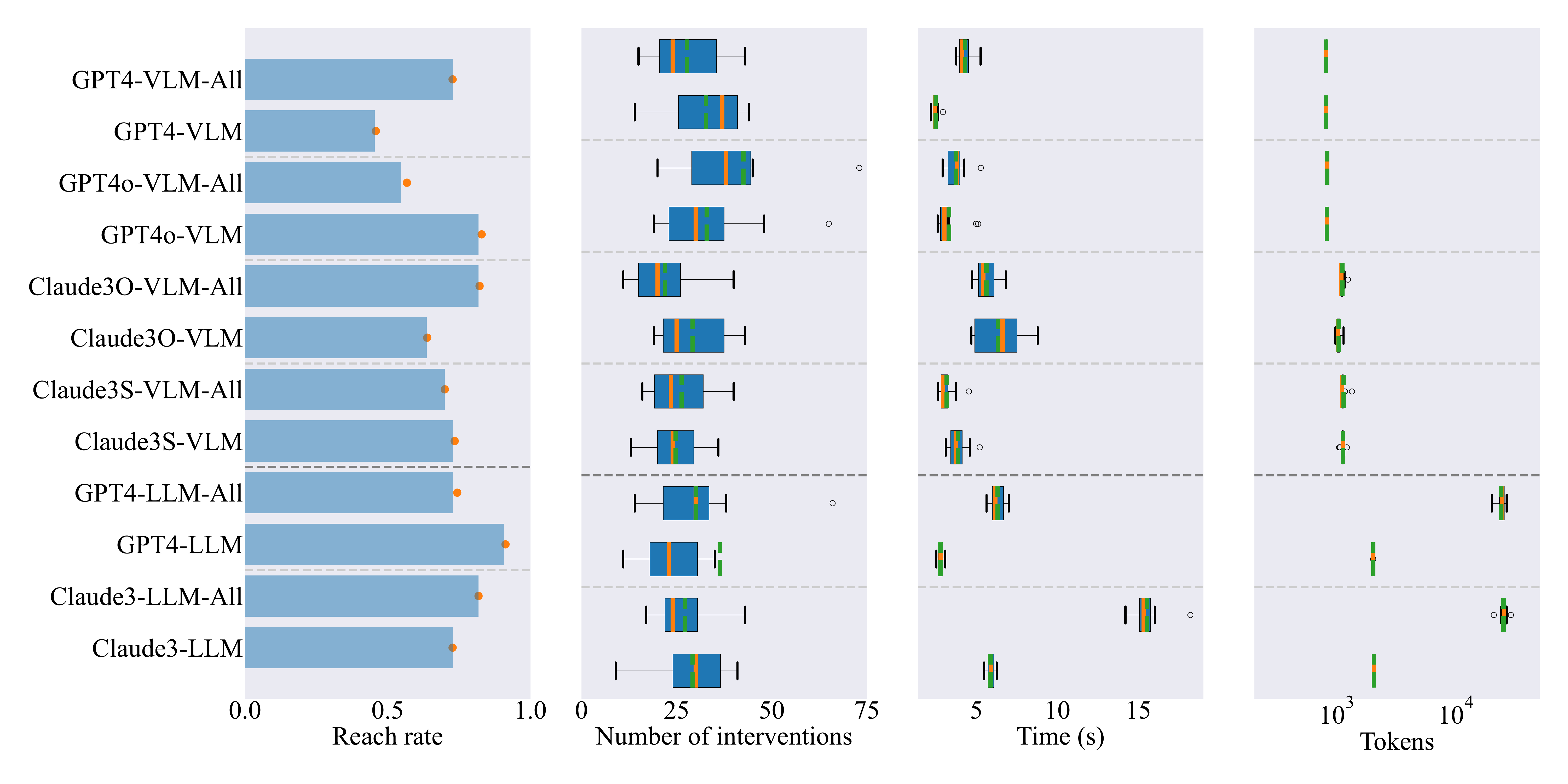}
    \caption{Performance of various high-level planners for  ``Maze'' environments with $N=50$ agents with all known environment information and partial information (the case with all known environment information is indicated with the suffix ``-All", e.g. ``GPT4-VLM-All"). 
    From left to right: 1) The bar shows the ratio of the trajectories where \textbf{all} the agents reach their goals over the total number of trajectories, and the orange dot shows the ratio of agents that reach their goals over all agents; 2) Box plot of the number of times the high-level planner intervened; 3) Box plot of the time spent for each high-level planner intervention; and 4) Box plot for the input + output token per intervention. In the box plots, the median values are in orange and the mean values are in green.}
    \label{fig:50 agents combine comp}
\end{figure*}

We evaluate the effect of providing partial environment information to the foundation models to study the trade-off between cost (in terms of the tokens) and performance. Figure \ref{fig:50 agents combine comp} compares the performance of querying a given foundation model with all collected observations of obstacles and querying it with only the most recent observations of obstacles ($50$ for LLMs and $100$ for VLMs). The rationale behind choosing the last few observations is twofold: 1) it reduces the number of tokens in the prompt provided to the foundation model, thereby reducing the time and cost per query, and 2)  in the considered environments, the initially observed obstacles do not play much role in determining the waypoints for the leader. 

\textbf{GPT4o-VLM and GPT4-LLM perform better with partial information}~ We can observe that, with the exception of \texttt{GPT-4o} VLM and \texttt{GPT4} LLM, the performance drops when partial information is used. It is not entirely clear why \texttt{GPT-4o} VLM and \texttt{GPT4} LLM perform better with partial information. Our understanding is that the performance of \texttt{GPT-4o} VLM and \texttt{GPT4} LLM is poor with the complete information due to their ability (or in this case, inability) of utilizing the provided information to make a good decision for this problem. 

\textbf{Smaller models perform better with partial information}~ From the results on \texttt{Claude3-Sonnet}-VLM, we can observe that the performance of a \textit{smaller} model (in terms of the model parameters) improves when partial information is used. On the other hand, for \textit{larger} models like \texttt{GPT4}-VLM and \texttt{Claude3}-LLM, the performance drops when only the partial information is used. It provides evidence in support of the intuition that smaller models do not perform well when more information is provided. 

\textbf{Quality of high-level commands deteriorates with less information}~ As discussed in the main paper, the number of high-level planner interventions is generally inversely proportional to the quality of the plan they suggest. As evident from the figure, the number of interventions with partial information is, on average, more than the case when all the known information is provided to the foundation models. We infer that the quality of the plan provided drops as the amount of data provided to the foundation models decreases. 

\textbf{Inference cost and time improves significantly}~ As expected, the average query time to foundation models (particularly LLMs) reduces significantly when using partial information. This provides a good trade-off metric for the user to determine how much data they should provide to the LLMs based on the desired level of performance and how much delay can be tolerated for a particular problem at hand. As stated in the beginning of the section, the motivation of conducting this ablation study is to see the cost-efficiency of providing partial information to the foundation models. While the average number of tokens used for VLMs does not change, there is a significant (at least an order of magnitude) reduction in the case of LLMs. Since the number of tokens is directly proportional to the cost associated with querying the proprietary foundation models, this trade-off study illustrates that an optimal amount of information can be provided to the foundation models to obtain desirable performance within a given cost and time budget. 

{
\subsection{Effect of multi-leader assignment}
Additionally, we evaluate the effect of using just one leader for large-scale MRS instead of the proposed multi-leader assignment as described in Section \ref{sec:multi leader assign}. 

\begin{figure*}[h]
    \centering
    \includegraphics[width=0.9\linewidth]{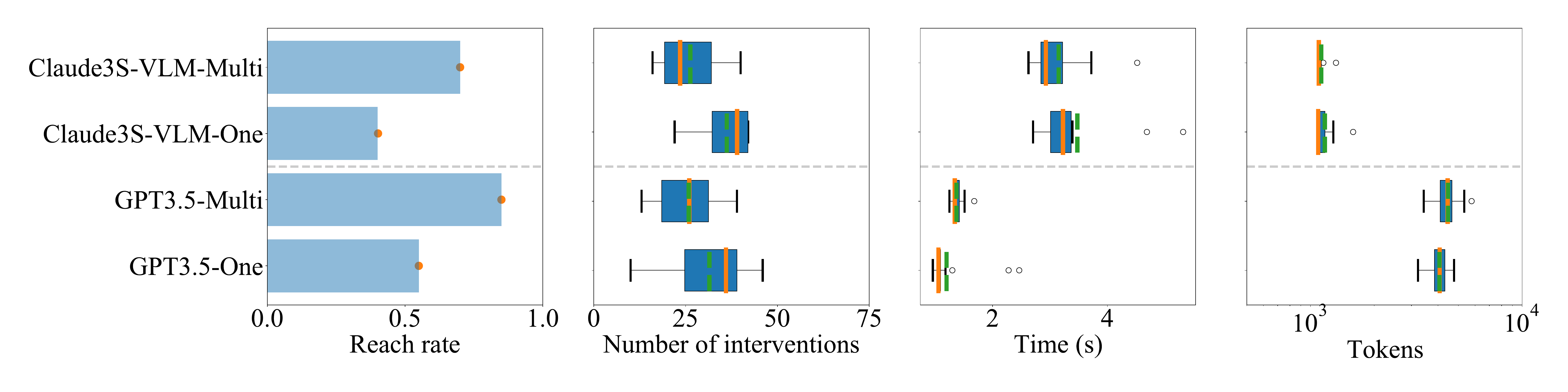}
    \caption{Performance of \texttt{Claude3-Sonnet-VLM} planner for ``Maze'' environments with $N=50$ agents and \texttt{GPT3.5-LLM} for ``Maze" environment with $N=25$ with a single leader and multi-leader assignment (the case with one leader is indicated with suffix ``-One", and that with multi-leader with ``-Multi''. 
    From left to right: 1) The bar shows the ratio of the trajectories where \textbf{all} the agents reach their goals over the total number of trajectories, and the orange dot shows the ratio of agents that reach their goals over all agents; 2) Box plot of the number of times the high-level planner intervened; 3) Box plot of the time spent for each high-level planner intervention; and 4) Box plot for the input + output token per intervention. In the box plots, the median values are in orange and the mean values are in green.}
    \label{fig:50 agents one multi leader}
\end{figure*}

As can be seen from the figure, the performance in terms of reach rate drops significantly when only one leader is used and all the other robots in MRS directly follow that leader as compared to the proposed multi-leader assignment where robots follow their local leaders. The number of interventions needed by the high-level planner is also higher when one leader is used instead of the proposed multi-leader framework. This illustrates the efficacy of the multi-leader assignment in large-scale systems. 
}

\section{Proof of no deadlocks}\label{app: proof of complete}

Recall that a deadlock is defined as the existence of an interval $\tau\subset \mathbb R$, given as $\tau =[a, b)$ where $a, b \in\mathbb R$, where the average speed of the MRS $\frac{1}{N}\sum_i|\dot p_i|(t) = 0$ for all $t\in \tau$ and the Lebesgue measure of the interval $\mu(\tau) > 0$, where $\mu(\tau) = |\tau| \coloneqq b-a$. Let $a\in \mathbb R$ be a time instant when the average speed of the MRS is zero, i.e., $\frac{1}{N}\sum_i|\dot p_i|(a) = 0$. Per the leader-assignment condition based on the average speed of the MRS, the condition $\frac{1}{N}\sum_i\|\dot p_i(t)\| < u_{min}$ triggers a leader assignment step at time $t$. The leader is chosen as the agent that has the minimum value of $\frac{\|p_i- p_{gi}\|}{\|\dot p_i|_{p_{gi, \mathrm{temp}}}\|}$. There are two cases possible: $\|\dot p_i|_{p_{gi, \mathrm{temp}}}\| = 0$ for all $i\in \mathcal V$ or there exists at least one $i\in \mathcal V$ such that $\|\dot p_i|_{p_{gi, \mathrm{temp}}}\| > 0$. The latter case leads to a non-zero average speed of the MRS, leading to $b = a$ in $\tau = [a, b)$, and thus, a deadlock cannot occur in this case. Next, we prove that $$\|\dot p_i|_{p_{gi, \mathrm{temp}}}\| = 0$$ for all $i\in \mathcal V$ is only possible at $t = 0$. Note that $\|\dot p_i|_{p_{gi, \mathrm{temp}}}\| = 0$ for all $i\in \mathcal V$ is possible only if \textit{all} the agents are occluded with obstacles and none of the agents have any free space to move. This is only possible at $a = 0$, i.e., the MRS is initialized in a location that is occluded by obstacles such that it cannot move. For $a > 0$, since the MRS can reach such a location where the average speed becomes lower than the leader-assignment threshold, there exists at least one agent that has \textit{free} space around it to move and hence, $\|\dot p_i|_{p_{gi, \mathrm{temp}}}\| = 0$ is not possible for all $i\in \mathcal V$ for any $a > 0$ and hence, the MRS cannot get stuck in a deadlock.

\end{document}